\documentclass[letterpaper]{article} 
\usepackage{aaai25}  
\usepackage{times}  
\usepackage{helvet}  
\usepackage{courier}  
\usepackage[hyphens]{url}  
\usepackage{graphicx} 
\usepackage{natbib}  
\usepackage{caption} 
\usepackage{algorithm}
\usepackage{algorithmic}

\usepackage{amsmath}
\usepackage{amssymb}
\usepackage{tabularray}
\usepackage{array}
\usepackage{booktabs}
\usepackage{vcell}

\usepackage{newfloat}
\usepackage{listings}
\DeclareCaptionStyle{ruled}{labelfont=normalfont,labelsep=colon,strut=off} 
\floatstyle{ruled}
\newfloat{listing}{tb}{lst}{}
\floatname{listing}{Listing}
\title{V2C-CBM: Building Concept Bottlenecks with Vision-to-Concept Tokenizer}
\author{
    Hangzhou He\textsuperscript{\rm 1,2,3,4}, Lei Zhu\textsuperscript{\rm 1,2,3,4}, Xinliang Zhang\textsuperscript{\rm 1,2,3,4}, Shuang Zeng\textsuperscript{\rm 1,2,3,4}, Qian Chen\textsuperscript{\rm 1,2,3,4}, Yanye Lu\textsuperscript{\rm 1,2,3,4}\thanks{Corresponding author}
}
\affiliations{
    \textsuperscript{\rm 1}Institute of Medical Technology, Peking University, Beijing, China\\
    \textsuperscript{\rm 2}Department of Biomedical Engineering, Peking University, Beijing, China\\
    \textsuperscript{\rm 3}National Biomedical Imaging Center, Peking University, Beijing, China\\
    \textsuperscript{\rm 4}Institute of Biomedical Engineering, Peking University Shenzhen Graduate School, Shenzhen, China\\

    \{zhuang, zhulei, chen\_qian\}@stu.pku.edu.cn, zhangxinliang\_mit@163.com, \{stevezs, yanye.lu\}@pku.edu.cn
}

\usepackage{bibentry}

\begin{document}

\maketitle

\begin{abstract}
Concept Bottleneck Models (CBMs) offer inherent interpretability by initially translating images into human-comprehensible concepts, followed by a linear combination of these concepts for classification. However, the annotation of concepts for visual recognition tasks requires extensive expert knowledge and labor, constraining the broad adoption of CBMs. Recent approaches have leveraged the knowledge of large language models to construct concept bottlenecks, with multimodal models like CLIP subsequently mapping image features into the concept feature space for classification. Despite this, the concepts produced by language models can be verbose and may introduce non-visual attributes, which hurts accuracy and interpretability. In this study, we investigate to avoid these issues by constructing CBMs directly from multimodal models. To this end, we adopt common words as base concept vocabulary and leverage auxiliary unlabeled images to construct a Vision-to-Concept (V2C) tokenizer that can explicitly quantize images into their most relevant visual concepts, thus creating a vision-oriented concept bottleneck tightly coupled with the multimodal model. This leads to our V2C-CBM which is training efficient and interpretable with high accuracy. Our V2C-CBM has matched or outperformed LLM-supervised CBMs on various visual classification benchmarks, validating the efficacy of our approach.
\end{abstract}

%
\begin{links}
     \link{Code}{https://github.com/riverback/V2C-CBM}
\end{links}

\section{Introduction}
With the increasing adoption of deep learning-based methods in high-risk and sensitive fields such as medical diagnosis and legal matters, the explainability of models is crucial for ensuring fairness and trustworthiness. Research is centered around two types of interpretability \cite{DBLP:journals/inffus/ArrietaRSBTBGGM20}: post-hoc and inherent. One benefit of post-hoc methods is that they do not hurt the performance of the original black-box models \cite{DBLP:journals/spm/NielsenDRRB22}. However, the fidelity of post-hoc methods cannot be guaranteed, and the explanations can be misleading \cite{pmlr-v235-geirhos24a} or unreliable without context \cite{pmlr-v235-tomaszewska24a}. In contrast, inherently interpretable models offer explainability through mechanism design, but their performance usually lags behind that of black-box deep learning models. The trade-off between accuracy and interpretability has been a focal point in the field of explainable artificial intelligence research \cite{gunning2019darpa,DBLP:journals/inffus/AliAEMACGSRH23}.

\begin{figure}[t]
    \centering
    \includegraphics[width=\linewidth]{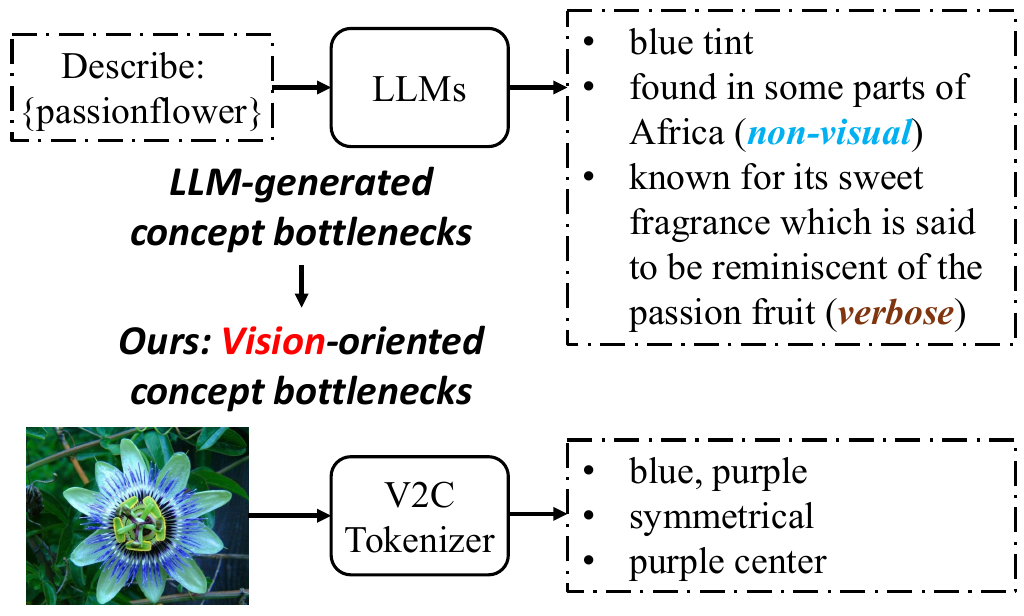}
    \caption{Problems in previous LLM-generated concept bottlenecks: non-visual and verbose concepts. Our solution: vision-oriented concept bottlenecks generated by Vision-to-Concept tokenizer directly from images.}
    \label{introduction_fig}
\end{figure}

Recently, concept bottleneck models (CBMs) have gained prominence for offering inherent interpretability with competitive performance \cite{koh2020concept}. CBMs first map the image features extracted by deep learning models into a set of human-interpretable concepts (such as \textit{red head} or \textit{white chest} for bird classification), and then employ a linear layer to aggregate these concepts for the final prediction. The two-step design of CBMs also allows for intervention by manually altering the concept predictions. When suitable and accurate concept labels are provided, CBM can achieve comparable accuracy with better interpretability. \citeauthor{koh2020concept} and \citeauthor{DBLP:conf/nips/ZarlengaBCMGDSP22} have demonstrated the effectiveness of CBMs in fine-grained bird classification \cite{wah2011caltech} and celebrity recognition \cite{liu2015deep} tasks.

Although CBMs hold promise, the annotation of concepts for visual recognition tasks requires considerable expert knowledge and labor, which impedes their widespread adoption and scalability. Recent research has addressed this challenge by using large language models (LLMs) to generate class-specific descriptions as concepts, and by harnessing pre-trained vision-language models (VLMs) to construct CBMs \cite{DBLP:conf/iccvw/PanousisIM23,DBLP:conf/iclr/MenonV23}. These methods have successfully scaled CBMs to datasets of the ImageNet scale, even attaining performance on par with the original VLMs. However, as illustrated in Figure \ref{introduction_fig}, the concepts generated by LLMs (such as GPT-3, \citeauthor{DBLP:conf/nips/BrownMRSKDNSSAA20}) are obtained by directly querying the LLM with class names, which presents two issues: 1) many of the generated concepts are \textbf{non-visual}, which are hard to be captured by the vision encoder, thereby reducing accuracy and faithfulness, 2) the concepts can be \textbf{verbose} which may contain multiple attributes in one concept, and it is hard to identify the exact concepts used by the model for prediction, diminishing the interpretability of CBMs. 

In this work, we propose to tackle these issues by directly generating class-specific concepts from images without the help of LLMs. To avoid verbose concepts, we use common words as our concept vocabulary and propose a concept filtering method to filter out non-visual attributes. Then a Vision-to-Concept (V2C) tokenizer is constructed using the vocabulary to quantize images into visual concepts. A contemporaneous work \cite{rao2024discover} also explores the idea of building concepts from common words, but their method requires an additional sparse autoencoder \cite{DBLP:conf/iclr/HubenCRES24} trained on the large-scale CC3M dataset with labels \cite{ng2020understanding} to name internal neurons as concepts. In contrast, our method can generate class-specific concepts without training, which is more efficient for resource-limited tasks like few-shot learning. We find that even without the knowledge of LLMs, our V2C tokenizer can still discover interpretable visual concepts. Our contributions can be summarized as follows.
\begin{enumerate}
    \item We propose the V2C tokenizer to discover visual concepts directly from images, avoiding the use of LLMs.
    \item We adopt common words as our concept vocabulary and develop a concept filtering method to remove non-visual and irrelevant concepts using auxiliary unlabeled images.
    \item The V2C-CBM that is built on the vision-oriented concepts generated by our V2C tokenizer can achieve high classification accuracy across various datasets with visually interpretable concepts.
\end{enumerate}

\section{Related Work}
\subsection{VLM-based Concept Bottleneck Models}
Traditional CBMs require annotated concepts for each class and training the concept predictor using these labels, which impedes the scalability of CBMs \cite{koh2020concept,DBLP:conf/icml/KimJPKY23,DBLP:conf/iclr/XuQMWL24}. Recent research has leveraged VLMs to project image features and concept texts into a shared feature space and use the cosine similarity as the concept predictor, making it more scalable for using numerous descriptions as concepts. LF-CBM is the first CBM that uses GPT-3 concepts and scales to ImageNet \cite{DBLP:conf/iclr/OikarinenDNW23}, and it removes concepts that are too long or similar and then uses CLIP-Dissect \cite{DBLP:conf/iclr/OikarinenW23} to filter out concepts that don't activate CLIP highly. \citeauthor{DBLP:conf/cvpr/YangPZJCY23} propose LaBo which harnesses GPT-3 to form class-specific bottlenecks and can be used for few-shot classification, and they also employ submodular optimization \cite{DBLP:journals/corr/abs-1010-4207} for concept selection. LM4CV proposes a learning-to-search method to discover a concise set of concepts generated by LLMs \cite{DBLP:conf/iccv/0003WZDHLWSM23}. Res-CBM translates black-box residual vectors with unclear meanings in PCBM-h \cite{DBLP:conf/iclr/YuksekgonulW023} into potential concepts to improve performance and preserve interpretability \cite{DBLP:journals/corr/abs-2404-08978}. Besides CBMs, some works also adopt a similar idea of leveraging language descriptions in improving the classification accuracy of VLMs, such as CDM \cite{DBLP:conf/iccvw/PanousisIM23} and DCLIP \cite{DBLP:conf/iclr/MenonV23}. However, all of the above methods require a set of concepts predefined by human experts or generated by LLMs. The former may reduce the scalability of CBMs since they need human effort and expert knowledge, while the latter has issues in that the concepts generated by LLMs may be overly verbose and non-visual.

\subsection{Image Quantization and Concept Discovery}
Image quantization methods aim to translate images into a set of discrete tokens from a codebook, or in the context of this paper, a set of concepts. A significant amount of work in this area is based on Auto-Encoder approaches, utilizing an encoder-decoder architecture to achieve image quantization through image reconstruction task \cite{DBLP:conf/nips/OordVK17,DBLP:conf/cvpr/EsserRO21,DBLP:conf/cvpr/LeeKKCH22,DBLP:conf/nips/ZarlengaBCMGDSP22,DBLP:conf/cvpr/HuangMCZ23,DBLP:conf/cvpr/ZhangZTL23,DBLP:conf/nips/LiuYA23,Zhu_2024_CVPR,zhu2024scalingcodebooksizevqgan}. The idea of using discrete codes to represent images features can also be used for extracting meaningful concepts from black-box models, which has shown to be successful in explaining LLMs \cite{DBLP:journals/corr/abs-2209-10652,DBLP:conf/iclr/HubenCRES24}. DN-CBM \cite{rao2024discover} adopts a similar idea of quantizing image features into the most similar concepts saved in the dictionary, then building a CBM using the discovered concepts. Yet their work requires additional training of a sparse autoencoder on a large dataset to discover these concepts. In contrast, we find that with the help of a large number of unlabeled web images, we can construct a V2C tokenizer and V2C-CBM directly using the VLMs, without the need for training or LLMs.

\section{Method}
\begin{figure*}[t]
    \centering
    \includegraphics[width=\linewidth]{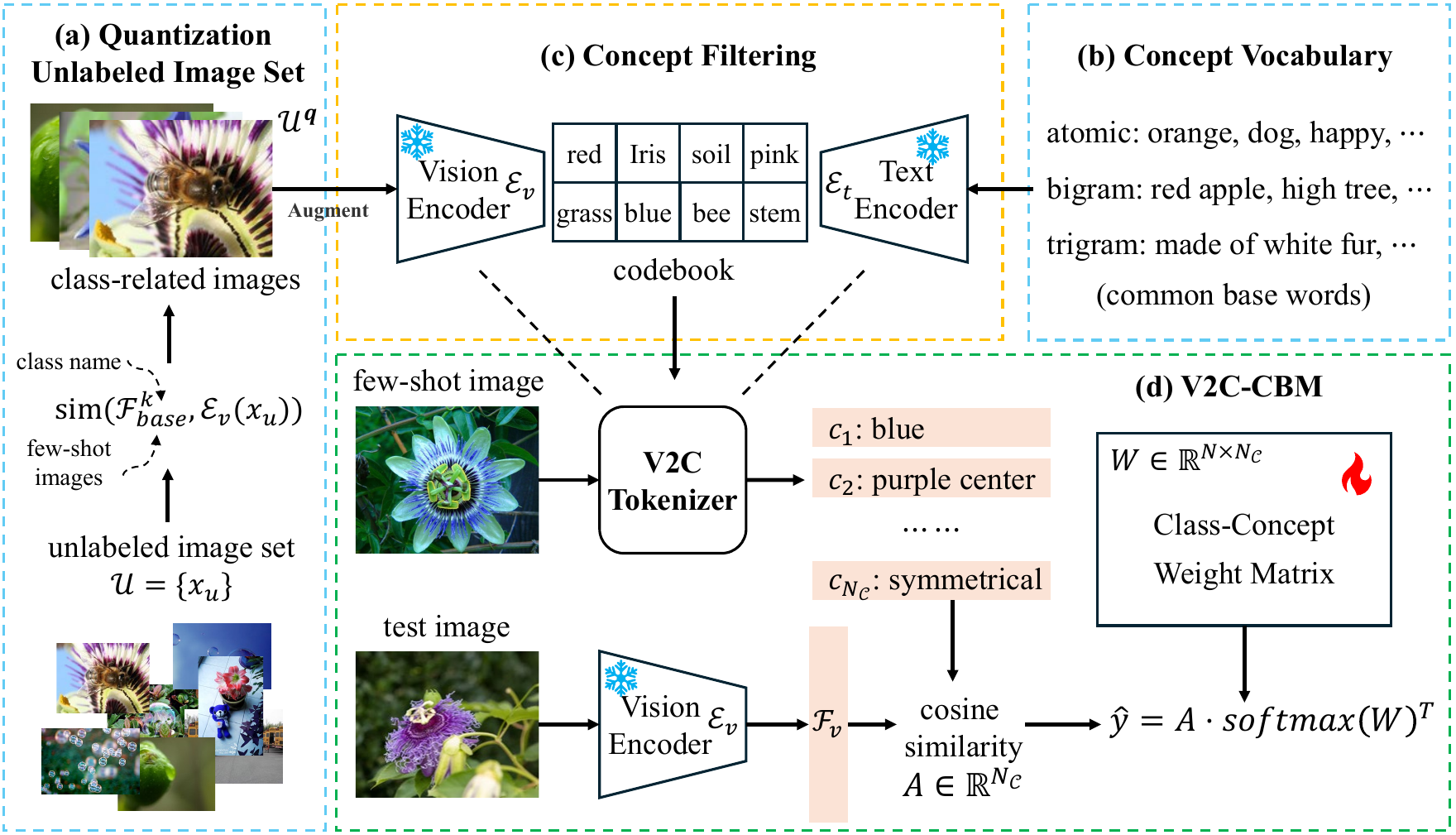}
    \caption{\textbf{Method overview}: (a) construct quantization unlabeled image set ${\cal U}^q$ using class-related base features ${\cal F}_{base}^k$ of class $k$, (b) adopt the most common words as the base concept vocabulary and use bigrams and trigrams to extend the vocabulary, (c) filter out non-visual and irrelevant concepts using ${\cal U}^q$ to form the codebook for V2C tokenizer, (d) build V2C-CBM with vision-oriented concept bottlenecks generated by V2C tokenizer from images. Our method is computation-efficient since we only need to train a class-concept weight matrix $W$, and don't require LLMs to discover class-specific concepts.}
    \label{v2c-tokenizer}
\end{figure*}

\subsection{Problem Definition and Method Overview}
Consider an image dataset ${\cal D}=\{(x, y)\}$ where $x$ is the image and $y\in {\cal Y}$ is a label from $N$ classes, and we have the class name or few-shot images $x_{fs}^k$ for each class $k$, we need a set of concepts of the dataset to build a CBM. 

In this work, instead of querying LLMs using class names to obtain the concept sets, we leverage an unlabeled image set ${\cal U}=\{x_u\}$ to construct a vision-to-concept (V2C) tokenizer ${\cal T}$, which can generate vision-oriented concepts directly from images. Figure \ref{v2c-tokenizer} presents an overview of our method. First, we use class-specific features ${\cal F}_{base}^k$ extracted from the class name or $x_{fs}$ of class $k$ to select unlabeled images, which forms the quantization unlabeled image set ${\cal U}^q$ (Figure \ref{v2c-tokenizer} (a)). Then, we use the most common words as our base concept vocabulary (Figure \ref{v2c-tokenizer} (b)) and propose a concept filtering method to use a VLM with vision encoder ${\cal E}_v$ and text encoder ${\cal E}_t$, and ${\cal U}^q$ to construct the codebook for V2C tokenizer (Figure \ref{v2c-tokenizer} (c)). And finally, we can build V2C-CBM using the concept bottlenecks generated by the V2C tokenizer from images (Figure \ref{v2c-tokenizer} (d)).

\subsection{Quantization Unlabeled Image Set}
We first build an unlabeled image set ${\cal U}=\{x_u\}$ to extract useful information for generating vision-oriented concepts, ${\cal U}$ can be obtained easily from large-scale web images. Then, we extract class-related features ${\cal F}_{base}^k$ for class $k$ from the class name ${\cal N}_k$ using text encoder ${\cal E}_t$ or few-shot images $x_{fs}^k$ using vision encoder ${\cal E}_v$:
\begin{align}
    \{{\cal F}_{base}^k\} &= \{{\cal E}_t({\cal P}({\cal N}_k) \} \quad \text{or}\\
    \{{\cal F}_{base}^k\} &= \{{\cal E}_v(x_{fs}^k)\}
\end{align}
where ${\cal P}$ is a set of predefined text prompts like ``a photo of \{class name\}'' to query the VLM to get more robust class features. Subsequently, we use the base features to quantize the unlabeled web images ${\cal U}$ into task-related ${\cal U}^q = \{x_u^q\}$. Specifically, we select the images with highest similarity scores from ${\cal U}$ for each class $k$ to form ${\cal U}^q = {\cal U}^q_1 \cup \cdots \cup {\cal U}^q_k \cup \cdots \cup {\cal U}^q_N$, where the subscript $k$ denotes the quantization dataset for the $k$-th class. The similarity is measured by the cosine similarity between the base class feature and the image feature:
\begin{equation}
    \text{sim}({\cal F}_{base}^k, {\cal E}_v(x_u)) = \frac{{\cal F}_{base}^k \cdot {\cal E}_v(x_u)}{\|{\cal F}_{base}^k\| \cdot \|{\cal E}_v(x_u)\|}
\end{equation}

\subsection{Concept Vocabulary}
Without the knowledge of LLMs, we simply adopt common words as the base concept vocabulary. Specifically, we first collect the most common words used to describe and indicate objects in daily life as the atomic vocabulary. To further enhance the representation ability of the vocabulary set, we include some relational vocabulary (such as \textit{part of}, \textit{made of}, \textit{is a}, and \textit{has a}), and use bigrams and trigrams to expand the vocabulary set. Because directly combining all the atomic words may lead to a large vocabulary and many unreasonable concepts, we construct the bigrams by combining adjectives $\{a\}$ and nouns $\{n\}$, and construct the trigrams by combining relational words $\{r\}$, adjectives and nouns. As exampled in Figure \ref{v2c-tokenizer} (b), the base concept vocabulary can be represented as:
\begin{equation*}
   {\cal C} = \{ {a},\cdots,{n_1},\cdots \} \cup \{ {a_1n_1},{a_1n_2},\cdots \} \cup \{{r_1a_1n_1},\cdots \}
\end{equation*}
We also remove concepts that contain class names in the dataset to prevent information leakage.

\subsection{Concept Filtering}
To filter out the non-visual and irrelevant vocabulary, we use the quantization unlabeled image set ${\cal U}^q$ with VLM to filter the concepts. Specifically, we use the text encoder ${\cal E}_t$ to extract the concept features ${\cal F}_{concept} = \{{\cal E}_t(c)\}$ for $c\in {\cal C}$. Then, we calculate the similarity between the concept feature and the image feature to filter out the non-visual (with low similarity) and irrelevant concepts (with low frequency). In order to better align the image features of unlabeled images with fine-grained concepts, for each image $x_u^q$, we also generate a set of augmented images ${\cal A}(x_u^q)$ to extend the image set. The augmented images are generated by applying random cropping, rotation, and then resizing to the original image. The similarity between the concept feature and the augmented image is calculated as:
\begin{equation}
    \text{sim}({\cal F}_{concept}^c, x_u^q) = \frac{{\cal F}_{concept}^c \cdot {\cal E}_v({\cal A}(x_u^q))}{\|{\cal F}_{concept}^c\| \cdot \|{\cal E}_v({\cal A}(x_u^q))\|}
\end{equation}
then we save $M$ most frequent concepts for each ${\cal U}_k^q$ as the final concept vocabulary ${\cal C} = {\cal C}_1 \cup {\cal C}_2 \cup \cdots \cup {\cal C}_N$, the size of the final concept vocabulary will be $N_{\cal C} = M \times N$, which will be used as the codebook of V2C tokenizer. Finally, we extract and save the concept features ${\cal F}_{concept} = \{{\cal E}_t(c)\}$ for each concept $c$ in the vocabulary as the embedding matrix.

\subsection{Vision-to-Concept Tokenizer}
With the concept vocabulary ${\cal C}$ and the saved features, we can construct the V2C tokenizer ${\cal T}$ to generate concepts by image quantization. Given an image $x$, the V2C tokenizer firstly extracts the image feature ${\cal F}_v$ using ${\cal E}_v$, then converts the image feature into top-$K$ nearest concepts depending on the Euclidean distance between the image feature and the saved concept features:
\begin{equation}
    {\cal T}({\cal F}_v) = \{c_1, c_2, \cdots, c_K\} = \mathop{\arg \min}\limits_{c\in {\cal C}}\|{\cal F}_v - {\cal E}_t(c)\|^2_2
\end{equation}
Given class-specific few-shot images, we can use the V2C tokenizer to generate the concepts for each class and select the most frequent concepts (depending on the size of the bottleneck) as the class-specific concept bottleneck.

\subsection{V2C-CBM}
Similar to other VLM-based CBMs, we use the vision encoder ${\cal E}_v$ and text encoder ${\cal E}_t$ to project the image $x$ and the set of concepts $c$ into a shared feature space and use cosine similarity scores $A$ as the concept prediction, then the final prediction is made by a linear layer optimized by images and image-level labels $y$:
\begin{equation}
    \hat{y} = \text{sim}({\cal E}_v(x), {\cal E}_t({\cal C}))
    \cdot \sigma(W)^T = A\cdot \sigma(W)^T
\end{equation}
\begin{equation}
    \mathop{\min}\limits_{W} {\cal L}(\hat{y}, y) = {\cal L}(A\cdot \sigma(W)^T, y)
\end{equation}
where $W\in \mathbb{R}^N \times \mathbb{R}^{N_{\cal C}}$ is the weight matrix of the linear layer, $\sigma$ is the softmax function applied along the concept axis, and ${\cal L}$ is the cross-entropy loss. Following \citeauthor{DBLP:conf/cvpr/YangPZJCY23}, we initialize the weight matrix $W$ with the concept priors of the V2C tokenizer to improve the few-shot classification performance when there is very little annotated data (e.g., 1- or 2-shots learning). Specifically, if a concept $c$ is generated by ${\cal T}$ using images of the $k$-th class, we set the corresponding elements of $W$ as 1, otherwise 0. For cases with more labeled images, we randomly initialize the weight matrix $W$ (more details in the ablation study section).

\section{Experimental Setup}

\subsection{Datasets}
We choose the following datasets for evaluation: CIFAR10, CIFAR100 \cite{2009Learning}, ImageNet \cite{DBLP:journals/ijcv/RussakovskyDSKS15} as the standard benchmarks for image classification; Aircraft \cite{maji13fine-grained}, CUB \cite{wah2011caltech}, Flower \cite{DBLP:conf/icvgip/NilsbackZ08}, and Food-101 \cite{DBLP:conf/eccv/BossardGG14} for fine-grained image classification; DTD \cite{DBLP:conf/cvpr/CimpoiMKMV14} for texture classification; RESISC45 \cite{DBLP:journals/pieee/ChengHL17} for remote sensing scene classification; and HAM10000 \cite{DBLP:journals/corr/abs-1803-10417} for skin tumor classification. We also use the same few-shot images and settings as LaBo and CLIP for a fair comparison. The classification accuracy on the test set is reported.

\subsection{Implementation Details}
We use CLIP ViT-L/14 to build our V2C tokenizer and V2C-CBM. For concept vocabulary, we use the English word frequency described in \cite{norvig2009natural}, and use NLTK library \cite{DBLP:journals/nle/Xue11} to determine adjectives and nouns to build the concept vocabulary. For the unlabeled image set ${\cal U}$, we randomly sample images from the ImageNet training set, and the default number of the unlabeled images is 200k. We use class names to extract the base features ${\cal F}_{base}$ for each class. $N_{\cal C}$ is set to 50 for all datasets. For each image, we select the top $K$ (set to 5) concepts to update frequency. We then rank the word frequencies and select the top $M$ (set to 500) words. Adam \cite{DBLP:journals/corr/KingmaB14} is used for optimization, and the detailed hyperparameters are provided in the supplementary material. All experiments are conducted on an NVIDIA A100 80G PCIE graphics card using PyTorch. Since our method only requires training the class-concept weight matrix $W$, it is computationally efficient.

\begin{table*}[t]
\centering
\footnotesize
\begin{tblr}{
  width = \linewidth,
  colspec = {Q[88]Q[55]Q[70]Q[70]Q[70]Q[65]Q[65]Q[65]Q[65]Q[65]Q[72]Q[80]},
  row{odd} = {c},
  row{4} = {c},
  row{6} = {c},
  row{8} = {c},
  cell{1}{1} = {r=2}{},
  cell{1}{2} = {r=2}{},
  cell{1}{3} = {c=10}{0.744\linewidth},
  cell{2}{4} = {c},
  cell{2}{5} = {c},
  cell{2}{6} = {c},
  cell{2}{7} = {c},
  cell{2}{8} = {c},
  cell{2}{9} = {c},
  cell{2}{10} = {c},
  cell{2}{11} = {c},
  cell{2}{12} = {c},
  hline{1,9} = {-}{0.08em},
  hline{3-4} = {-}{},
}
Model & Concept & CLIP ViT-L/14 &  &  &  &  &  &  &  &  & \\
 &  & Aircraft & CIFAR10 & CIFAR100 & CUB & DTD & Flower & Food & HAM & RESISC45 & ImageNet\\
LP & - & 64.0{ $\pm$0.28} & 98.0{ $\pm$0.02} & 87.5{ $\pm$0.08} & 84.5{ $\pm$0.10} & 81.5{ $\pm$0.39} & 99.5{ $\pm$0.01} & 93.2{ $\pm$0.09} & 82.9{ $\pm$0.36} & 93.9{ $\pm$0.82} & 83.9{ $\pm$0.09}\\
LaBo & GPT-3 & \textbf{61.3}{ $\pm$0.22} & 97.8{ $\pm$0.11} & 86.0{ $\pm$0.02} & 81.9{ $\pm$0.03} & 76.9{ $\pm$0.10} & \textbf{99.3}{ $\pm$0.01} & 92.4{ $\pm$0.05} & 80.8{ $\pm$0.55} & 91.1{ $\pm$0.78} & 84.0{ $\pm$0.06}\\
CDM & GPT-3 & - & 98.0 & \textbf{86.4} & - & - & - & - & - & - & 83.4\\
DCLIP & GPT-3 & - & - & - & 63.5 & 54.4 & - & 92.4 & - & - & 75.0\\
DN-CBM & SAE & - & \textbf{98.1} & 86.0 & - & - & - & - & - & - & 83.6\\
Ours & - & 60.7{ $\pm$0.01} & 98.0{ $\pm$0.03} & \textbf{86.4}{ $\pm$0.01} & \textbf{83.0}{ $\pm$0.12} & \textbf{78.2}{ $\pm$0.29} & 98.8{ $\pm$0.14} & \textbf{92.8}{ $\pm$0.04} & \textbf{81.0}{ $\pm$0.12} & \textbf{92.6}{ $\pm$0.26} & \textbf{84.1}{ $\pm$0.05}
\end{tblr}
\caption{Classification accuracy (\%). LP stands for linear probing. The results of DCLIP and DN-CBM  are from their respective works, and CDM is from the DN-CBM paper. The standard deviation is derived from three random experiments. \label{classification_performance_full_shot}}
\end{table*}

\begin{figure*}[t]
    \centering
    \includegraphics[width=\linewidth]{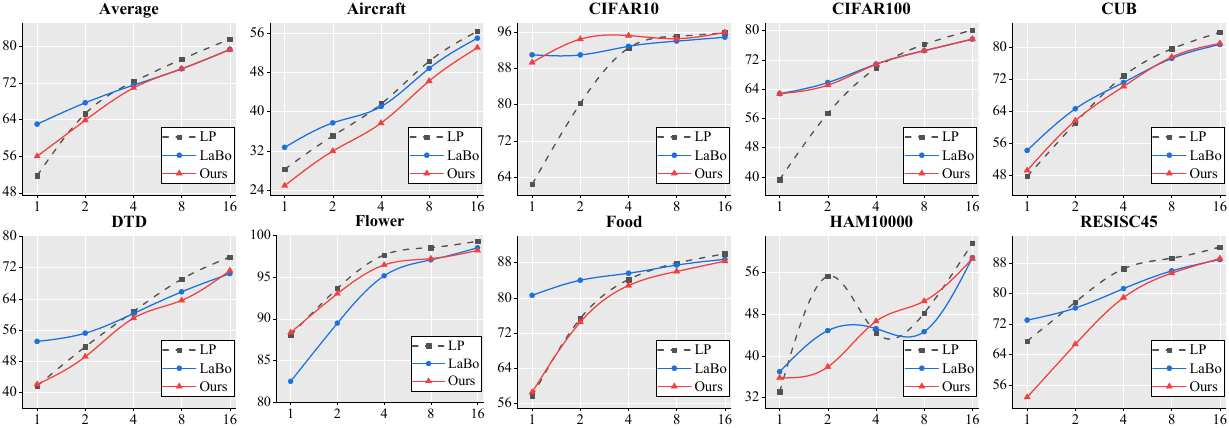}
    \caption{Few-shot classification accuracy (x-axis denotes the number of shots and y-axis denotes test accuracy).}
    \label{few-shot_test_set_fig}
\end{figure*}

\begin{table}[t]
\centering
\footnotesize
\begin{tblr}{
  width = \linewidth,
  colspec = {Q[202]Q[115]Q[115]Q[115]Q[115]Q[115]Q[115]},
  cells = {c},
  cell{1}{1} = {r=2}{},
  cell{1}{2} = {c=6}{0.69\linewidth},
  hline{1,6} = {-}{0.08em},
  hline{3,4} = {-}{},
}
Method & number of shots &  &  &  &  & \\
 & 1 & 2 & 4 & 8 & 16 & all\\
LP & 51.8 & 65.3 & 72.3 & 77.1 & 81.6 & 86.9\\
LaBo & \textbf{63.0} & \textbf{67.7} & \textbf{71.5} & 75.1 & 79.3 & 85.2\\
Ours & 57.8 & 64.0 & 71.1 & \textbf{75.8} & \textbf{79.7} & \textbf{85.6}
\end{tblr}
\caption{Average classification accuracy (\%) on all datasets.\label{average_classification_accuracy}}
\end{table}

\begin{table*}[t]
\centering
\footnotesize
\begin{tabular}{>{\centering\hspace{0pt}}m{0.14\linewidth}|>{\hspace{0pt}}m{0.12\linewidth}|>{\hspace{0pt}}m{0.24\linewidth}|>{\hspace{0pt}}m{0.14\linewidth}|>{\hspace{0pt}}m{0.20\linewidth}} 
\toprule
\textbf{Class Name} & \multicolumn{1}{>{\centering\hspace{0pt}}m{0.12\linewidth}|}{\textbf{V2C Tokenizer}} & \multicolumn{1}{>{\centering\hspace{0pt}}m{0.24\linewidth}|}{\textbf{LaBo}} & \multicolumn{1}{>{\centering\hspace{0pt}}m{0.14\linewidth}|}{\textbf{CDM}} & \multicolumn{1}{>{\centering\arraybackslash\hspace{0pt}}m{0.20\linewidth}}{\textbf{DCLIP}} \\ 
\hline
\includegraphics[width=2cm]{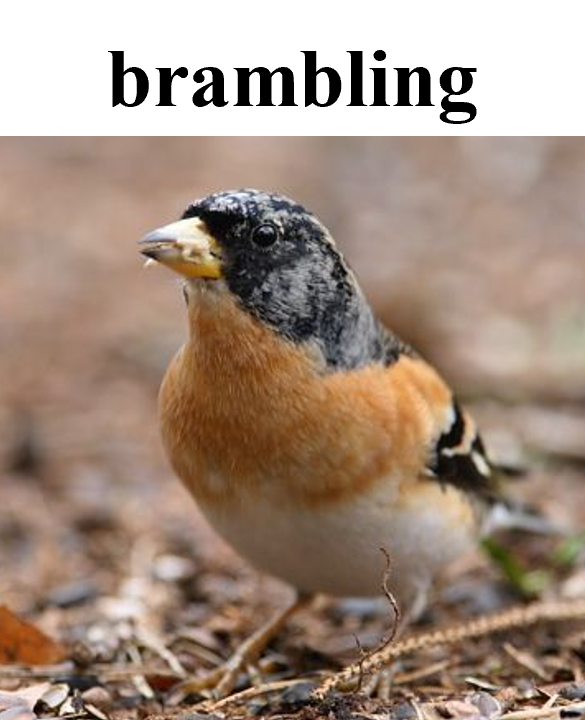} & \textbf{1.} black head\par{}\textbf{2.} brown back\par{}\textbf{3.} common bird & \textbf{1.} small, sparrow-like bird with a streaked brown back\par{}\textbf{2.} found in woods and forests across Europe and Asia\par{}\textbf{3.} found in woods and forests in Europe and Asia & \textbf{1.} barrow\par{}\textbf{2.} a long, thin, orange root\par{}\textbf{3.} large wings & \textbf{1.} a small, sparrow-like bird\par{}\textbf{2.} brown and white plumage\par{}\textbf{3.} a black head with a white stripe above the eye \\ 
\hline
\includegraphics[width=2cm]{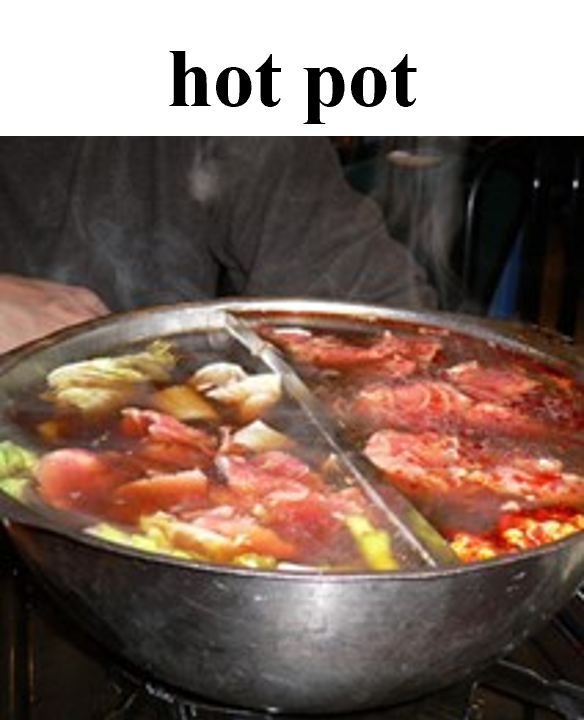} & \textbf{1.} hot bowl\par{}\textbf{2.} hot dishes\par{}\textbf{3.} red soup & \textbf{1.} circular metal object with a handle on the side\par{}\textbf{2.} round metal container with a handle on the side\par{}\textbf{3.} conical lid with a knob in the center & \textbf{1.} droopy lips and ears\par{}\textbf{2.} a game room\par{}\textbf{3.} a small, pointed tail & \textbf{1.} a pot or other container with a heating element\par{}\textbf{2.} a power cord\par{}\textbf{3.} a bowl or other vessel for holding food \\ 
\hline
\includegraphics[width=2cm]{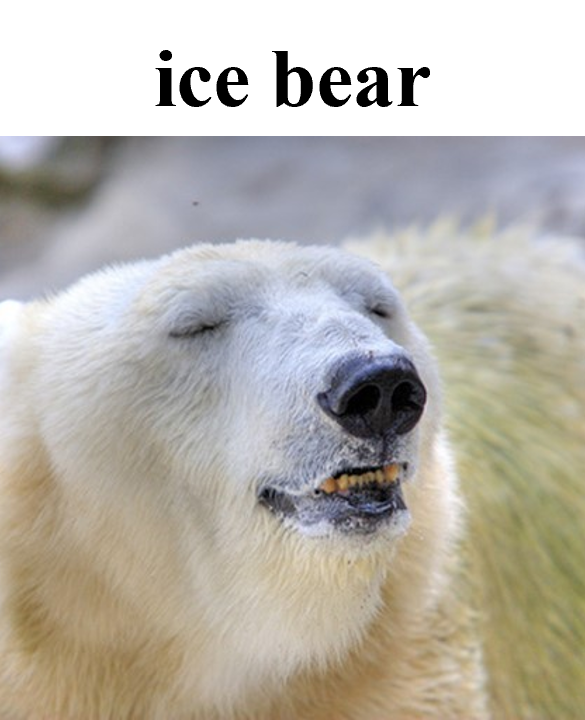} & \textbf{1.} white bear\par{}\textbf{2.} white enclosure\par{}\textbf{3.} white animal & \textbf{1.} large, white bear that lives in the Arctic\par{}\textbf{2.} perfect for keeping the bear warm in its icy habitat\par{}\textbf{3.} very important bear & \textbf{1.} white wingtips\par{}\textbf{2.} referees\par{}\textbf{3.} a note from Santa & \textbf{1.} large, white bear\par{}\textbf{2.} long neck\par{}\textbf{3.} small ears \\
\bottomrule
\end{tabular}
\caption{Top-3 concept Visualization of different methods.\label{concept_visualization_on_imagenet}}
\end{table*}

\section{Evaluation}
\subsection{Baselines}
We compare the classification performance of our V2C-CBM with other concept label-free methods including LaBo \cite{DBLP:conf/cvpr/YangPZJCY23}, CDM \cite{DBLP:conf/iccvw/PanousisIM23}, DCLIP \cite{DBLP:conf/iclr/MenonV23} and DN-CBM \cite{rao2024discover}, and compare the few-shot classification performance with LaBo. The concept sets are kept the same as their original settings for a fair comparison. We report the results using the same backbone for all methods, and the linear probe (LP) performance of the black-box model is also provided for reference. 
\subsection{Classification Accuracy}
In Table \ref{classification_performance_full_shot}, we present the classification accuracy of our V2C-CBM on ten datasets. Our method achieves classification accuracy that is better than or comparable to the baseline methods across all datasets, without leveraging the prior knowledge of LLMs like LaBo and CDM, or additional training of a SAE for concept discovery like DN-CBM. V2C-CBM also surpasses the black-box linear probing performance of CLIP on the ImageNet dataset, showing the great scalability of our method on large datasets. Although lacking the extensive knowledge provided by LLMs or Internet encyclopedias (such as Wikipedia and WordNet), we show that making full use of the unlabeled images can also lead to competitive or even better performance, and can discover class-specific interpretable visual concepts with our method (more details in Table \ref{concept_visualization_on_imagenet}).
\subsection{Few-shot Classification Performance}
Figure \ref{few-shot_test_set_fig} illustrates the few-shot performance of our V2C-CBM on 9 datasets. Due to the lack of extensive prior knowledge possessed by LLMs, our V2C-CBM usually performs less effectively than the GPT-3 concepts guided method LaBo when the number of labeled images is very small (1-shot and 2-shot), especially in fine-grained classification tasks like aircraft and food classification, but it can achieve comparable or slightly superior performance to the black-box linear probing method with few labeled images on almost all datasets with better interpretability. 

However, the classification accuracy of V2C-CBM can increase quickly as the number of labeled images grows. As shown in Table \ref{average_classification_accuracy}, V2C-CBM achieves accuracy close to LaBo in 4-shots learning and exceeds it after 4-shots. We think this makes sense because the more labeled images, the more robust and accurate the concepts generated by the V2C tokenizer from images will be. For example, the \textit{black head} concepts might not be generated by our V2C tokenizer when this part of the objects happens to be obscured in the limited few-shot images. In contrast, the advantage of LaBo lies in its ability to generate robust concept descriptions for similar categories with the help of the prior knowledge of LLMs, thus enhancing the generalization of the model. 

\subsection{Concept Visualization}
To see whether the V2C tokenizer can discover valid visual concepts without the help of LLMs, we iterate through all the images of a particular class in the datasets and select the most frequent concepts used by our V2C tokenizer. The visualization of the top-3 concepts for some classes in the ImageNet dataset is illustrated in Table \ref{concept_visualization_on_imagenet}. Compared to other methods, our V2C tokenizer can generate concise and visually informative concepts, capturing the salient features of the target classes. For example, the \textit{white} for ice bear and the \textit{black head} for brambling birds. We provide more concept visualization in the supplementary material.

\section{Ablation Study}
\subsection{Size of the Unlabeled Image Set}
In Table \ref{ablation_on_size_of_unlabeled_imageset}, we investigate the impact of the size of the unlabeled image set on the final performance of V2C-CBM. As the size of $\cal U$ increases from 1k to 200k, the model's performance shows an overall increasing trend. Since a $\cal U$ of size 200k has proven to be sufficiently effective and considering the runtime efficiency, we don't continue to increase the dataset size and use 200k as our default setting. 

\begin{table}[t]
\centering
\footnotesize
\begin{tblr}{
  width = \linewidth,
  colspec = {Q[219]Q[71]Q[135]Q[110]Q[123]Q[123]Q[123]},
  cells = {c},
  cell{1}{1} = {r=2}{},
  cell{1}{2} = {c=6}{0.685\linewidth},
  hline{1,7} = {-}{0.08em},
  hline{3} = {-}{},
}
Dataset & number of unlabeled images &  &  &  &  & \\
 & 1k & 40k & 80k & 120k & 160k & 200k\\
CIFAR10 & 97.6 & 97.7 & 97.8 & 97.9 & 97.5 & \textbf{98.0}\\
CUB & 80.3 & 81.4 & 81.6 & 81.9 & 82.2 & \textbf{83.0}\\
DTD & 73.1 & 76.3 & 76.8 & 77.4 & 77.6 & \textbf{78.0}\\
RESISC45 & 90.2 & 91.8 & 91.9 & 92.0 & 92.0 & \textbf{92.6}
\end{tblr}
\caption{Classification accuracy (\%) using unlabeled image sets with different number of images.\label{ablation_on_size_of_unlabeled_imageset}}
\end{table}

\subsection{Bigram and Trigram Concepts}
We also investigate the impact of different combinations of conceptual vocabulary on model performance. Specifically, we explore the scenarios of using only atomic concepts (A), using both atomic and bigram concepts (AB), and employing atomic, bigram, and trigram concepts simultaneously (ABT). The experimental results are presented in Table \ref{ablation_on_bi-trigram_concepts}. We find that the model achieves better performance in the ABT scenario. Looking back at the discovered concepts shown in Table \ref{concept_visualization_on_imagenet}, we think the combination is beneficial for describing similar objects with more accurate and distinct attributes (e.g., \textit{white fur} v.s. \textit{fur}). This may also explain why LaBo can exhibit stronger few-shot capabilities in fine-grained classification tasks besides the knowledge provided by LLMs — while concepts generated by LLMs are more complex, assigning diverse concepts to distinguish similar categories becomes easier. However, when the number of labeled images is sufficiently large, the advantage of using complex concepts is no longer pronounced, particularly in the context of our vision-oriented concept bottleneck.

\begin{table}[t]
\centering
\footnotesize
\begin{tblr}{
  width = \linewidth,
  colspec = {Q[200]Q[110]Q[100]Q[100]Q[100]Q[100]Q[100]Q[100]},
  cells = {c},
  cell{1}{1} = {r=2}{},
  cell{1}{2} = {r=2}{},
  cell{1}{3} = {c=6}{0.6\linewidth},
  cell{3}{1} = {r=3}{},
  cell{6}{1} = {r=3}{},
  hline{1,9} = {-}{0.08em},
  hline{3,6} = {-}{},
}
Dataset & Type & number of shots &  &  &  &  & \\
 &  & 1 & 2 & 4 & 8 & 16 & all\\
Food & A & 52.3 & 73.2 & 81.7 & 85.7 & 88.0 & 92.0\\
 & AB & 57.7 & 73.9 & 82.0 & 85.8 & 88.0 & 92.1\\
 & ABT & \textbf{58.6} & \textbf{74.5} & \textbf{82.5} & \textbf{86.3} & \textbf{88.9} & \textbf{92.8}\\
{RESISC45} & A & 60.5 & 68.2 & 78.4 & 84.7 & 88.4 & 91.6\\
 & AB & 61.9 & 69.8 & 78.1 & 84.3 & 88.4 & 91.5\\
 & ABT & \textbf{62.0} & \textbf{69.8} & \textbf{80.7} & \textbf{86.6} & \textbf{89.7} & \textbf{92.6}
\end{tblr}
\caption{Classification accuracy (\%) using different concept combination method: A denotes atomic words, B denotes bigrams and T denotes trigrams.\label{ablation_on_bi-trigram_concepts}}
\end{table}

\begin{table}[t]
\centering
\footnotesize
\begin{tblr}{
  width = \linewidth,
  colspec = {Q[279]Q[124]Q[124]Q[124]Q[124]Q[124]},
  cells = {c},
  cell{1}{1} = {r=2}{},
  cell{1}{2} = {c=6}{0.624\linewidth},
  hline{1,5} = {-}{0.08em},
  hline{3,4} = {-}{},
}
{Base Feature\\Type} & number of shots &  &  &  &  & \\
  & 1 & 2 & 4 & 8 & 16 \\
class name & \textbf{88.4} & \textbf{93.0} & \textbf{96.5} & \textbf{97.2} & \textbf{98.2} \\
{few-shot\\images} & {77.2\\{(11.2↓)}} & {86.6\\{(6.4↓)}} & {93.8\\{(2.7↓)}} & {96.4\\{(0.8↓)}} & {97.7\\{(0.5↓)}}  
\end{tblr}
\caption{Classification accuracy (\%) on the Flower dataset using different class-related base features.\label{ablation_study_on_different_base_features}}
\end{table}

\begin{table}[t]
\centering
\footnotesize
\begin{tblr}{
  width = \linewidth,
  colspec = {Q[200]Q[120]Q[120]Q[120]Q[120]Q[120]Q[120]},
  cells = {c},
  cell{1}{1} = {r=2}{},
  cell{1}{2} = {c=6}{0.648\linewidth},
  hline{1,6} = {-}{0.08em},
  hline{3-5} = {-}{},
}
{Vocabulary\\Set} & number of shots &  &  &  &  & \\
 & 1 & 2 & 4 & 8 & 16 & all\\
base & \textbf{58.6} & \textbf{74.5} & \textbf{82.5} & \textbf{86.3} & \textbf{88.9} & \textbf{92.8}\\
{without\\food} & {29.7\\{ (28.9↓)}} & {58.3\\{ (16.2↓)}} & {74.5\\{ (8.0↓)}} & {80.3\\{ (6.0↓)}} & {83.2\\{ (5.7↓)}} & {86.3\\{ (6.5↓)}}\\
{only\\food} & {50.2\\{ (8.4↓)}} & {71.9\\{ (2.6↓)}} & {81.2\\{ (1.4↓)}} & {85.3\\{ (1.0↓)}} & {87.7\\{ (1.2↓)}} & {92.4\\{ (0.4↓)}}
\end{tblr}
\caption{Classification accuracy (\%) on the Food dataset using different vocabulary sets.\label{ablation_on_different_vocabulary_set}}
\end{table}

\subsection{Different Types of Class-related Features}
We also examine the discrepancy in extracting class-related features ${\cal F}_{base}^k$ by employing class names versus using few-shot images. For class names, we utilize a collection of 85 prompt templates, leveraging the text encoder to extract features and computing the mean of these features to serve as the base feature ${\cal F}_{base}$. The specifics of the text prompts are detailed in the supplementary material; for few-shot images, we compute the mean of the extracted image features as the base feature. The performance on the Flower-102 dataset is illustrated in Table \ref{ablation_study_on_different_base_features}. We observe that the use of class names and text prompts demonstrates better robustness, particularly when the number of labeled images is limited. The gap between the text features derived from class names and the image features diminishes as the number of labeled images increases. This is consistent with our previous understanding of the few-shot performance, which is the more images, the more robust the concepts generated by the V2C tokenizer.

\subsection{Different Vocabulary Set}
We also test the impact of different sources of concept vocabulary on model performance. In Table \ref{ablation_on_different_vocabulary_set}, we try, respectively: 1) using the top 10k most common English words (base); 2) removing words that can be used to describe food in the base vocabulary, then using the rest of the words as another vocabulary, which produces a vocabulary unrelated to the classification task (without food); and 3) using the filtered out words as the vocabulary (only food). It can be observed that when using vocabulary unrelated to the task (without food), the model’s performance significantly drops, with an accuracy reduction of up to 28.9\% in 1-shot tasks and 16.2\% in 2-shot tasks. The base vocabulary achieves similar but higher accuracy than the only food vocabulary, which indicates that our method can effectively discover relevant concepts to the target task from the large vocabulary.

\begin{table}[t]
\centering
\footnotesize
\begin{tblr}{
  width = \linewidth,
  colspec = {Q[298]Q[102]Q[102]Q[102]Q[102]Q[102]Q[102]},
  cells = {c},
  cell{1}{1} = {r=2}{},
  cell{1}{2} = {c=6}{0.612\linewidth},
  hline{1,7} = {-}{0.08em},
  hline{3-5} = {-}{},
}
Method & number of shots &  &  &  &  & \\
 & 1 & 2 & 4 & 8 & 16 & all\\
LP & 51.8 & 65.3 & 72.3 & 77.1 & 81.6 & 87.0\\
LaBo & 63.0 & 67.7 & 71.5 & 75.1 & 79.3 & 85.3\\
Ours${}_p$ & \textbf{57.8} & \textbf{64.0} & 70.3 & 74.7 & 78.7 & 85.2\\
Ours${}_r$ & 50.4 & 61.6 & \textbf{71.1} & \textbf{75.8} & \textbf{79.7} & \textbf{85.7}
\end{tblr}
\caption{Average classification accuracy (\%) on all datasets using different initialization methods. Ours${}_p$ means initializing with concept priors and Ours${}_r$ means random.\label{ablation_study_on_different_initialization_methods}}
\end{table}

\subsection{Initialization with Priors}
Following \citet{DBLP:conf/cvpr/YangPZJCY23}, we also test whether using concept priors in the V2C tokenizer (the language priors in LaBo) can improve the classification performance. As shown in Table \ref{ablation_study_on_different_initialization_methods}, initializing with concept priors is more effective for very limited labeled images (1- and 2-shots), yet random initialization is more powerful as the number of labeled images grows. The conclusion is similar to LaBo's, which states that prior is more important for low shot settings since there is less signal to guide concept importance. So we use concept priors to initialize $W$ for 1- and 2-shot learning while using random initialization for the others.

\section{Conclusion}
In this work, we construct a vision-oriented concept bottleneck without the reliance on LLMs by developing a V2C tokenizer that maps images to a discrete set of concepts. The creation of the V2C tokenizer necessitates only a set of unlabeled images, which can be readily acquired from the Internet. We show that the V2C tokenizer can discover interpretable visual concepts and lead to V2C-CBM, which has surpassed LLM-guided CBMs across various datasets and even outperformed black-box linear probing methods on the ImageNet dataset, showcasing the efficacy of our method. Follow-up work may be devoted to developing methods for evaluating the trustworthiness of VLM-based CBMs with open-vocabulary concepts.

\appendix

\section{Acknowledgments}
This work was supported in part by the Natural Science Foundation of China under Grants 62394314, 82371112, 623B2001, and 62394311 and in part by the High-grade, Precision and Advanced University Discipline Construction Project of Beijing (BMU2024GJJXK004).

\bibliography{aaai25}
\newpage
\section{Supplemental Materials}
\vspace{1cm}
\section{Dataset Statistics}
Table \ref{detailed_statistics} presents the detailed statistics for all datasets. We use the same train/val/test splits provided by \citet{DBLP:conf/cvpr/YangPZJCY23} for fair comparison. The random images selected for few-shot learning are saved after the first experiment, and are used consistently in subsequent experiments. 
\vspace{0.5cm}
\section{Text Prompts}
In order to extract class-related base features ${\cal F}_{base}^k$ for class $k$ from the class name ${\cal N}_k$, we use a set of text prompts ${\cal P}$ to extract more robust features. We list five prompts in Table \ref{text_prompts_table} and all the 85 prompts are provided in the code appendix. In addition, we add a superclass name like \textit{aricraft} for the Aircraft dataset and \textit{texture} for the DTD dataset, shown in Table \ref{text_prompts_table_aircraft} and \ref{text_prompts_table_dtd}.
\vspace{0.5cm}
\section{Detailed Results}
The full numerical results on the test set for all few-shot experiments are shown in Table \ref{full_numerical_results}. The detailed results on all datasets for the ablation study on initialization priors are provided in Table \ref{appendix_on_full_results_for_initialization_prios}. In the main experimental results, we choose initialization with priors for 1- and 2- shots, while choosing random initialization with the others.
\vspace{0.5cm}
\section{Different Vocabulary Sizes}
We also study the impact of different vocabulary sizes on the final model performance. The experimental results are shown in Table \ref{appendix_ablation_on_different_vocabulary_size}. The results indicate that our method is not highly sensitive to the size of the vocabulary. Therefore, to reduce runtime, we can opt for a relatively smaller vocabulary. Additionally, when the vocabulary size reaches 20k or 30k, it already encompasses a significant number of uncommon words, yet there is no marked change in model performance. This also suggests that our method is robust to the noise present in larger vocabularies.
\vspace{0.5cm}
\section{Hyperparameters}
We list all the hyperparameters that we set for the experiments in Table \ref{appendix_hyperparameters}. All the hyperparameters are obtained by 5 grid search runs and tuned on the val sets. The parameters with the highest val accuracy are the final hyperparameters.

\begin{table}[t]
\centering
\begin{tblr}{
  width = \linewidth,
  colspec = {Q[248]Q[129]Q[213]Q[154]Q[154]},
  cells = {c},
  cell{1}{1} = {r=2}{},
  cell{1}{3} = {c=3}{0.521\linewidth},
  hline{1,13} = {-}{0.08em},
  hline{2} = {3-5}{},
  hline{3} = {-}{},
}
\textbf{Name} & \textbf{n. of} & \textbf{n. of images} &  & \\
 & \textbf{class} & train & val & test\\
Aircraft & 102 & 3,334 & 3,333 & 3,333\\
CIFAR-10 & 10 & 45,000 & 5,000 & 10,000\\
CIFAR-100 & 100 & 45,000 & 5,000 & 5,000\\
CUB & 200 & 3,994 & 2,000 & 5,794\\
DTD & 47 & 2,820 & 1,128 & 1,692\\
Flower & 102 & 4,093 & 1,633 & 2,463\\
Food & 101 & 50,500 & 20,200 & 30,300\\
HAM10000 & 7 & 8,010 & 1,000 & 1,005\\
RESISC45 & 45 & 3,150 & 3,150 & 25,200\\
ImageNet & 1,000 & 1,281,167 & 50,000 & -
\end{tblr}
\caption{Detailed statistics for all datasets.\label{detailed_statistics}}
\end{table}

\begin{table}[t]
\centering
\begin{tblr}{
  width = 0.8\linewidth,
  colspec = {Q[885]},
  row{1} = {c},
  hline{1,8} = {1}{0.08em},
  hline{2} = {1}{},
}
Base Prompt Templates\\
1. a photo of a \{${\cal N}_k$\}.\\
2. a jpeg corupted photo of a \{${\cal N}_k$\}.\\
3. a photo of a large \{${\cal N}_k$\}.\\
4. a toy \{${\cal N}_k$\}.\\
5. is a type of \{${\cal N}_k$\}.\\
$\cdots \cdots$\\
\end{tblr}
\caption{Text prompts used for extracting base features.\label{text_prompts_table}}
\end{table}

\begin{table}[t]
\centering
\begin{tblr}{
  width = 0.8\linewidth,
  colspec = {Q[885]},
  row{1} = {c},
  hline{1,8} = {1}{0.08em},
  hline{2} = {1}{},
}
Prompt Templates\\
1. a photo of a \{${\cal N}_k$\} aircraft.\\
2. a jpeg corupted photo of a \{${\cal N}_k$\} aircraft.\\
3. a photo of a large \{${\cal N}_k$\} aircraft.\\
4. a toy \{${\cal N}_k$\} aircraft.\\
5. is a type of \{${\cal N}_k$\} aircraft.\\
$\cdots \cdots$\\
\end{tblr}
\caption{Text prompts used for Aircraft dataset.\label{text_prompts_table_aircraft}}
\end{table}

\begin{table}[t]
\centering
\begin{tblr}{
  width = 0.8\linewidth,
  colspec = {Q[885]},
  row{1} = {c},
  hline{1,8} = {1}{0.08em},
  hline{2} = {1}{},
}
Prompt Templates for DTD\\
1. a photo of a \{${\cal N}_k$\} texture.\\
2. a jpeg corupted photo of a \{${\cal N}_k$\} texture.\\
3. a photo of a large \{${\cal N}_k$\} texture.\\
4. a toy \{${\cal N}_k$\} texture.\\
5. is a type of \{${\cal N}_k$\} texture.\\
$\cdots \cdots$\\
\end{tblr}
\caption{Text prompts used for DTD dataset.\label{text_prompts_table_dtd}}
\end{table}

\begin{table}[t]
\centering
\begin{tblr}{
  width = \linewidth,
  colspec = {Q[233]Q[171]Q[98]Q[98]Q[98]Q[98]Q[98]},
  cells = {c},
  cell{1}{1} = {r=2}{},
  cell{1}{2} = {r=2}{},
  cell{1}{3} = {c=5}{0.49\linewidth},
  vline{2} = {1-32}{},
  hline{1,33} = {-}{0.08em},
  hline{3,6,9,12,15,18,21,24,27,30} = {-}{},
  hline{4,7,10,13,16,19,22,25,28,31} = {2-7}{},
}
Dataset & Method & number of shots &  &  &  & \\
 &  & 1 & 2 & 4 & 8 & 16\\
 & LP & 51.8 & 65.3 & 72.3 & 77.1 & 81.6\\
Average & LaBo & \textbf{63.0} & \textbf{67.7} & \textbf{71.5} & 75.1 & 79.3\\
 & Ours & 58.1 & 64.1 & 71.1 & \textbf{75.8} & \textbf{79.7}\\
 & LP & 28.3 & 35.1 & 41.6 & 50.3 & 56.4\\
Aircraft & LaBo & \textbf{32.7} & \textbf{37.7} & \textbf{41.0} & \textbf{48.8} & \textbf{55.0}\\
 & Ours & 24.9 & 32.0 & 38.5 & 48.0 & 53.5\\
 & LP & 62.4 & 80.3 & 92.5 & 95.1 & 95.9\\
CIFAR10 & LaBo & \textbf{91.0} & 91.0 & 92.9 & 94.1 & 94.9\\
 & Ours & 89.3 & \textbf{94.5} & \textbf{94.9} & \textbf{95.7} & \textbf{96.0}\\
 & LP & 39.3 & 57.4 & 69.7 & 76.2 & 80.2\\
CIFAR100 & LaBo & 62.7 & \textbf{65.8} & \textbf{70.8} & 74.5 & 77.7\\
 & Ours & \textbf{62.7} & 65.1 & 69.5 & \textbf{74.9} & \textbf{78.7}\\
 & LP & 41.7 & 51.7 & 60.8 & 69.0 & 74.7\\
CUB & LaBo & \textbf{54.2} & \textbf{64.6} & 71.2 & 77.2 & 80.7\\
 & Ours & 49.2 & 61.7 & \textbf{71.4} & \textbf{78.4} & \textbf{82.2}\\
 & LP & 88.1 & 93.7 & 97.7 & 98.6 & 99.3\\
DTD & LaBo & \textbf{53.1} & \textbf{55.2} & \textbf{60.2} & \textbf{65.8} & 70.5\\
 & Ours & 42.1 & 49.2 & 59.4 & 64.3 & \textbf{70.9}\\
 & LP & 47.7 & 61.1 & 72.8 & 79.6 & 83.7\\
Flower & LaBo & 82.5 & 89.5 & 95.2 & 97.1 & 98.5\\
 & Ours & \textbf{88.4} & \textbf{93.0} & \textbf{96.5} & \textbf{97.2} & \textbf{98.2}\\
 & LP & 57.8 & 75.3 & 84.2 & 87.9 & 90.0\\
Food & LaBo & \textbf{80.6} & \textbf{84.0} & \textbf{85.6} & \textbf{87.5} & 88.8\\
 & Ours & 58.6 & 74.5 & 82.5 & 86.3 & \textbf{88.9}\\
 & LP & 33.1 & 55.3 & 44.5 & 48.3 & 61.7\\
{\small HAM10000} & LaBo & 37.0 & \textbf{44.9} & 45.3 & 44.7 & \textbf{58.9}\\
 & Ours & \textbf{45.8} & 37.3 & \textbf{46.7} & \textbf{50.6} & 58.7\\
 & LP & 67.6 & 77.8 & 86.5 & 89.3 & 92.2\\
RESISC45 & LaBo & \textbf{73.1} & \textbf{76.2} & \textbf{81.3} & 85.9 & 88.9\\
 & Ours & 62.0 & 69.8 & 80.7 & \textbf{86.6} & \textbf{89.7}
\end{tblr}
\caption{Detailed results for the few-shot experiments.\label{full_numerical_results}}
\end{table}

\begin{table}[t]
\centering
\begin{tblr}{
  width = \linewidth,
  colspec = {Q[250]Q[100]Q[90]Q[90]Q[90]Q[90]Q[90]Q[90]},
  row{1} = {c},
  column{4} = {c},
  column{5} = {c},
  column{6} = {c},
  column{7} = {c},
  cell{1}{1} = {r=2}{},
  cell{1}{2} = {r=2}{},
  cell{1}{3} = {c=6}{0.539\linewidth},
  cell{2}{3} = {c},
  cell{3}{1} = {r=2}{c},
  cell{3}{2} = {c},
  cell{3}{3} = {c},
  cell{4}{2} = {c},
  cell{4}{3} = {c},
  cell{5}{1} = {r=2}{c},
  cell{5}{2} = {c},
  cell{5}{3} = {c},
  cell{6}{2} = {c},
  cell{6}{3} = {c},
  cell{7}{1} = {r=2}{c},
  cell{7}{2} = {c},
  cell{7}{3} = {c},
  cell{8}{2} = {c},
  cell{8}{3} = {c},
  cell{9}{1} = {r=2}{c},
  cell{9}{2} = {c},
  cell{9}{3} = {c},
  cell{10}{2} = {c},
  cell{10}{3} = {c},
  cell{11}{1} = {r=2}{c},
  cell{11}{2} = {c},
  cell{11}{3} = {c},
  cell{12}{2} = {c},
  cell{12}{3} = {c},
  cell{13}{1} = {r=2}{c},
  cell{13}{2} = {c},
  cell{13}{3} = {c},
  cell{14}{2} = {c},
  cell{14}{3} = {c},
  cell{15}{1} = {r=2}{c},
  cell{15}{2} = {c},
  cell{15}{3} = {c},
  cell{16}{2} = {c},
  cell{16}{3} = {c},
  cell{17}{1} = {r=2}{c},
  cell{17}{2} = {c},
  cell{17}{3} = {c},
  cell{18}{2} = {c},
  cell{18}{3} = {c},
  cell{19}{1} = {r=2}{c},
  cell{19}{2} = {c},
  cell{19}{3} = {c},
  cell{20}{2} = {c},
  cell{20}{3} = {c},
  cell{21}{1} = {r=2}{c},
  cell{21}{2} = {c},
  cell{21}{3} = {c},
  cell{22}{2} = {c},
  cell{22}{3} = {c},
  vline{2} = {1-22}{},
  hline{1,23} = {-}{0.08em},
  hline{3,5,7,9,11,13,15,17,19,21} = {-}{},
}
Dataset & Method & number of shots &  &  &  &  & \\
 &  & 1 & 2 & 4 & 8 & 16 & all\\
Average & Ours${}_p$ & \textbf{57.8} & \textbf{64.0} & 70.3 & 74.7 & 78.7 & 85.2\\
 & Ours${}_r$ & 50.4 & 61.6 & \textbf{71.1} & \textbf{75.8} & \textbf{79.7} & \textbf{85.7}\\
Aircraft & Ours${}_p$ & \textbf{24.9} & 32.0 & 37.7 & 46.2 & 53.0 & 60.3\\
 & Ours${}_r$ & 24.9 & \textbf{32.5} & \textbf{38.5} & \textbf{48.0} & \textbf{53.5} & \textbf{60.7}\\
{CIFAR10} & Ours${}_p$ & \textbf{89.3} & \textbf{94.5} & \textbf{95.3} & 94.6 & 96.0 & 97.9\\
 & Ours${}_r$ & 79.6 & 91.5 & 94.9 & \textbf{95.7} & \textbf{96.0} & \textbf{98.0}\\
{CIFAR100} & Ours${}_p$ & \textbf{62.7} & \textbf{65.1} & \textbf{70.8} & 74.5 & 77.7 & 86.3\\
 & Ours${}_r$ & 41.5 & 58.5 & 69.5 & \textbf{74.9} & \textbf{78.7} & \textbf{86.4}\\
CUB & Ours${}_p$ & \textbf{49.2} & 61.7 & 70.2 & 77.6 & 81.0 & 82.1\\
 & Ours${}_r$ & 48.9 & \textbf{62.1} & \textbf{71.4} & \textbf{78.4} & \textbf{82.2} & \textbf{83.0}\\
DTD & Ours${}_p$ & \textbf{42.1} & \textbf{49.2} & 59.0 & 63.6 & \textbf{71.2} & 78.0\\
 & Ours${}_r$ & 37.6 & 45.7 & \textbf{59.4} & \textbf{64.3} & 70.9 & \textbf{78.5}\\
Flower & Ours${}_p$ & \textbf{88.4} & \textbf{93.0} & 92.7 & 95.1 & 97.3 & \textbf{99.0}\\
 & Ours${}_r$ & 79.5 & 86.3 & \textbf{96.5} & \textbf{97.2} & \textbf{98.2} & 98.7\\
Food & Ours${}_p$ & \textbf{58.6} & \textbf{74.5} & \textbf{82.8} & 86.0 & 88.4 & 92.1\\
 & Ours${}_r$ & 52.5 & 73.3 & 82.5 & \textbf{86.3} & \textbf{88.9} & \textbf{92.8}\\
{\small HAM10000} & Ours${}_p$ & \textbf{45.8} & 37.3 & 45.1 & 49.5 & 54.9 & 78.8\\
 & Ours${}_r$ & 35.8 & \textbf{37.9} & \textbf{46.7} & \textbf{50.6} & \textbf{58.7} & \textbf{81.0}\\
{RESISC45} & Ours${}_p$ & \textbf{59.6} & \textbf{69.2} & 79.0 & 85.4 & 89.2 & 92.1\\
 & Ours${}_r$ & 52.9 & 66.7 & \textbf{80.7} & \textbf{86.6} & \textbf{89.7} & \textbf{92.6}
\end{tblr}
\caption{Detailed numerical results for initialization with priors. Ours${}_p$ means initialization with concept priors and Ours${}_r$ means random.\label{appendix_on_full_results_for_initialization_prios}}
\end{table}

\begin{table}[t]
\centering
\begin{tblr}{
  width = \linewidth,
  colspec = {Q[169]Q[121]Q[121]Q[121]Q[121]Q[121]Q[121]},
  cells = {c},
  cell{1}{1} = {r=2}{},
  cell{1}{2} = {c=6}{0.726\linewidth},
  hline{1,9} = {-}{0.08em},
  hline{3-8} = {-}{},
  vline{2} = {1-9}{},
}
{Vocab.\\Size} & number of shots &  &  &  &  & \\
 & 1 & 2 & 4 & 8 & 16 & all\\
3k & 58.9 & 66.7 & 79.5 & 84.8 & 89.2 & 92.5\\
5k & 62.7 & 70.1 & 79.7 & 85.4 & 89.3 & 92.7\\
10k & 59.6 & 69.2 & 80.7 & 86.6 & 89.7 & 92.6\\
15k & 63.3 & 68.7 & 79.2 & 84.5 & 89.1 & 92.5\\
20k & 60.7 & 68.8 & 79.4 & 84.3 & 89.2 & 92.5\\
30k & 62.6 & 67.9 & 79.4 & 84.0 & 89.1 & 92.4
\end{tblr}
\caption{Classification accuracy (\%) on the RESISC45 dataset using different sizes of the vocabulary set.\label{appendix_ablation_on_different_vocabulary_size}}
\end{table}

\begin{table*}[t]
\centering
\footnotesize
\begin{tblr}{
  width = \linewidth,
  colspec = {Q[98]Q[148]Q[115]Q[115]Q[115]Q[115]Q[115]Q[115]},
  cells = {c},
  cell{1}{1} = {r=2}{},
  cell{1}{2} = {r=2}{},
  cell{1}{3} = {c=6}{0.69\linewidth},
  cell{3}{1} = {r=4}{},
  cell{7}{1} = {r=4}{},
  cell{11}{1} = {r=4}{},
  cell{15}{1} = {r=4}{},
  cell{19}{1} = {r=4}{},
  cell{23}{1} = {r=4}{},
  cell{27}{1} = {r=4}{},
  cell{31}{1} = {r=4}{},
  cell{35}{1} = {r=4}{},
  cell{39}{1} = {r=4}{},
  vline{2-3} = {1-43}{},
  vline{3} = {1-43}{},
  hline{1,43} = {-}{0.08em},
  hline{3,7,11,15,19,23,27,31,35,39} = {-}{},
}
Dataset & Parameter & number of shots &  &  &  &  & \\
 &  & 1 & 2 & 4 & 8 & 16 & all\\
Aircraft & K & 50 & 50 & 50 & 50 & 50 & 50\\
 & Learning Rate & $5e^{-6}$ & $5e^{-6}$ & $5e^{-6}$ & $5e^{-6}$ & $5e^{-6}$ & $5e^{-6}$\\
 & Batch Size & 16 & 32 & 64 & 128 & 256 & 256\\
 & Max Epochs & 10,000 & 10,000 & 10,000 & 10,000 & 10,000 & 10,000\\
CIFAR10 & K & 50 & 50 & 50 & 50 & 50 & 50\\
 & Learning Rate & $1e^{-4}$ & $1e^{-4}$ & $1e^{-4}$ & $1e^{-4}$ & $1e^{-4}$ & $1e^{-4}$\\
 & Batch Size & 2 & 4 & 8 & 16 & 32 & 512\\
 & Max Epochs & 15,000 & 15,000 & 15,000 & 15,000 & 15,000 & 15,000\\
CIFAR100 & K & 50 & 50 & 50 & 50 & 50 & 50\\
 & Learning Rate & $1e^{-5}$ & $1e^{-5}$ & $1e^{-5}$ & $1e^{-5}$ & $1e^{-5}$ & $1e^{-5}$\\
 & Batch Size & 16 & 32 & 64 & 128 & 256 & 512\\
 & Max Epochs & 10,000 & 10,000 & 10,000 & 10,000 & 10,000 & 10,000\\
CUB & K & 50 & 50 & 50 & 50 & 50 & 50\\
 & Learning Rate & $5e^{-5}$ & $5e^{-5}$ & $5e^{-5}$ & $5e^{-5}$ & $5e^{-5}$ & $5e^{-5}$\\
 & Batch Size & 32 & 64 & 128 & 256 & 512 & 512\\
 & Max Epochs & 5,000 & 5,000 & 5,000 & 5,000 & 5,000 & 5,000\\
DTD & K & 50 & 50 & 50 & 50 & 50 & 50\\
 & Learning Rate & $1e^{-5}$ & $1e^{-5}$ & $1e^{-5}$ & $1e^{-5}$ & $5e^{-5}$ & $1e^{-4}$\\
 & Batch Size & 8 & 16 & 32 & 64 & 256 & 512\\
 & Max Epochs & 15,000 & 15,000 & 15,000 & 15,000 & 15,000 & 15,000\\
Flower & K & 25 & 25 & 25 & 25 & 25 & 25\\
 & Learning Rate & $1e^{-5}$ & $1e^{-5}$ & $1e^{-5}$ & $1e^{-5}$ & $1e^{-5}$ & $5e^{-5}$\\
 & Batch Size & 8 & 16 & 32 & 64 & 256 & 256\\
 & Max Epochs & 20,000 & 20,000 & 20,000 & 20,000 & 20,000 & 20,000\\
Food & K & 50 & 50 & 50 & 50 & 50 & 50\\
 & Learning Rate & $1e^{-5}$ & $1e^{-4}$ & $1e^{-4}$ & $1e^{-4}$ & $1e^{-4}$ & $1e^{-5}$\\
 & Batch Size & 16 & 32 & 64 & 128 & 256 & 1,024\\
 & Max Epochs & 5,000 & 5,000 & 5,000 & 5,000 & 5,000 & 5,000\\
HAM10000 & K & 50 & 50 & 50 & 50 & 50 & 50\\
 & Learning Rate & $1e^{-3}$ & $1e^{-3}$ & $1e^{-4}$ & $1e^{-3}$ & $1e^{-3}$ & $5e^{-4}$\\
 & Batch Size & 4 & 4 & 8 & 8 & 16 & 256\\
 & Max Epochs & 10,000 & 10,000 & 10,000 & 10,000 & 10,000 & 10,000\\
RESISC45 & K & 50 & 50 & 50 & 50 & 50 & 50\\
 & Learning Rate & $5e^{-5}$ & $5e^{-5}$ & $5e^{-5}$ & $5e^{-5}$ & $5e^{-5}$ & $5e^{-5}$\\
 & Batch Size & 8 & 16 & 32 & 64 & 128 & 256\\
 & Max Epochs & 15,000 & 15,000 & 15,000 & 15,000 & 15,000 & 15,000\\
ImageNet & K & - & - & - & - & - & 50\\
 & Learning Rate & - & - & - & - & - & $1e^{-5}$\\
 & Batch Size & - & - & - & - & - & 1,024\\
 & Max Epochs & - & - & - & - & - & 1,000
\end{tblr}
\caption{All hyperparameters used for the main experiments. $K$ is the number of concepts per class. \label{appendix_hyperparameters}}
\end{table*}
\end{document}